%% file: main.tex
\documentclass[10pt,twocolumn,letterpaper]{article}

\usepackage[pagenumbers]{cvpr} 

\usepackage{colortbl}
\usepackage{color}
\usepackage{CJKutf8}
\usepackage{pifont}
\usepackage{tcolorbox}
\usepackage{gensymb}
\definecolor{cvprblue}{rgb}{0.21,0.49,0.74}
\usepackage[pagebackref,breaklinks,colorlinks,allcolors=cvprblue]{hyperref}
\usepackage{multirow} 
\def\paperID{1875} 
\def\confName{CVPR}
\def\confYear{2025}
\title{Beyond the Destination: A Novel Benchmark for Exploration-Aware \\Embodied Question Answering}

\author{
Kaixuan Jiang\textsuperscript{1}\quad Yang Liu\textsuperscript{1}\footnotemark[1]\quad
Weixing Chen\textsuperscript{1}\quad
 Jingzhou Luo \textsuperscript{1}\quad 
 Ziliang Chen \textsuperscript{2}\quad 
 Ling Pan \textsuperscript{3}\quad\\ 
 Guanbin Li\textsuperscript{1,2}\quad Liang Lin\textsuperscript{1,2}\\
{\normalsize
\textsuperscript{1}Sun Yat-sen University \quad
\textsuperscript{2}Peng Cheng Laboratory \quad
\textsuperscript{3} Hong Kong University of Science and Technology  \quad
}\\
\tt\small jiangkx3@mail2.sysu.edu.cn, 
liuy856@mail.sysu.edu.cn, chenwx228@mail2.sysu.edu.cn, \\
\tt\small luojzh5@mail2.sysu.edu.cn, 
c.ziliang@yahoo.com,
penny.ling.pan@gmail.com
\\
\tt\small 
liguanbin@mail.sysu.edu.cn,
linliang@ieee.org
\\
\normalsize \color{red}{https://github.com/HCPLab-SYSU/EXPRESS-Bench}
}
\begin{document}

\twocolumn[{%
\renewcommand\twocolumn[1][]{#1}%
\maketitle
\centering
\includegraphics[width=1\textwidth]{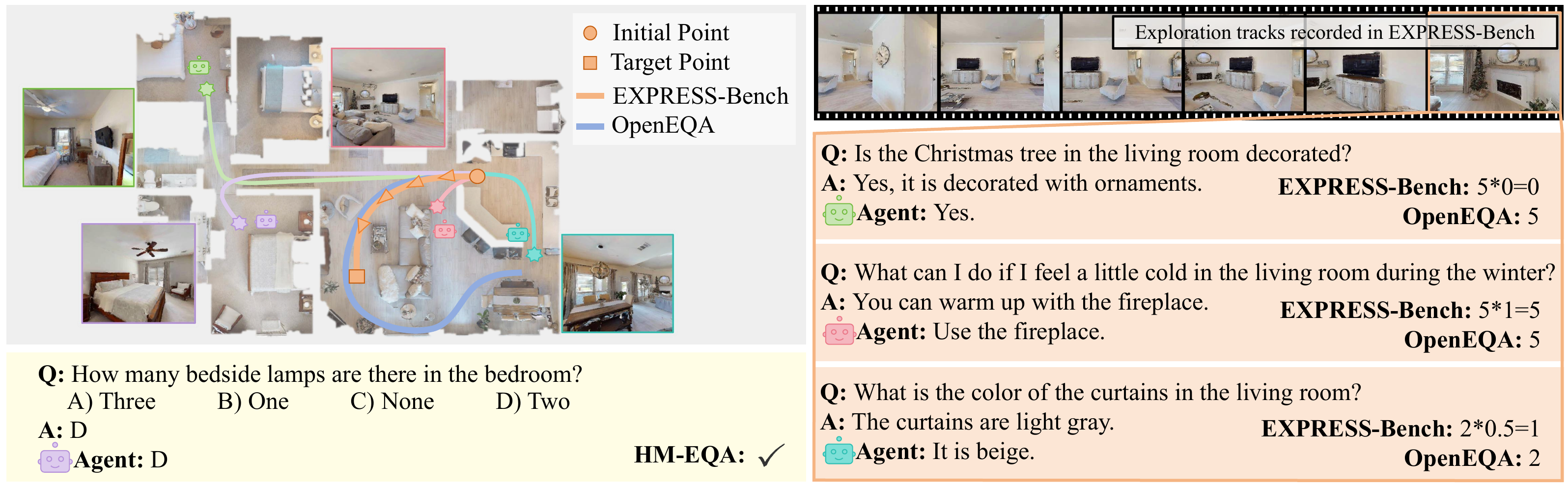}
\captionof{figure}{Comparison of our EXPRESS-Bench with other EQA datasets. 
The orange trajectory in the top-down map shows a complete exploration path from EXPRESS-Bench, with observation images at key waypoints (top-right). Data for this path is in the orange box. The blue trajectory simulates OpenEQA's episodic memory, passing near the target but not ending there. The yellow box simulates how multiple-choice data is generated in HM-EQA, lacking the exploration path. For each question, answers are based on visual observations at the endpoint, scored according to each dataset’s evaluation method. Unlike HM-EQA and OpenEQA, which may give higher scores based on answer similarity, EXPRESS-Bench adjusts scores for incorrect or fabricated answers by grounding them in the agent’s observations.
}
}]
\renewcommand{\thefootnote}{\fnsymbol{footnote}}
\footnotetext[1]{Corresponding author}

\begin{abstract}
Embodied Question Answering (EQA) is a challenging task in embodied intelligence that requires agents to dynamically explore 3D environments, actively gather visual information, and perform multi-step reasoning to answer questions. 
However, current EQA approaches suffer from critical limitations in exploration efficiency, dataset design, and evaluation metrics. Moreover, existing datasets often introduce biases or prior knowledge, leading to disembodied reasoning, while frontier-based exploration strategies struggle in cluttered environments and fail to ensure fine-grained exploration of task-relevant areas.
To address these challenges, we construct the \textbf{EXP}loration-awa\textbf{R}e \textbf{E}mbodied que\textbf{S}tion an\textbf{S}wering \textbf{Bench}mark (EXPRESS-Bench), the largest dataset designed specifically to evaluate both exploration and reasoning capabilities. EXPRESS-Bench consists of 777 exploration trajectories and 2,044 question-trajectory pairs.
To improve exploration efficiency, we propose Fine-EQA, a hybrid exploration model that integrates frontier-based and goal-oriented navigation to guide agents toward task-relevant regions more effectively. Additionally, we introduce a novel evaluation metric, Exploration-Answer Consistency (EAC), which ensures faithful assessment by measuring the alignment between answer grounding and exploration reliability. Extensive experimental comparisons with state-of-the-art EQA models demonstrate the effectiveness of our EXPRESS-Bench in advancing embodied exploration and question reasoning.
\end{abstract}

\begin{table*}[t]
\centering\scriptsize
\caption{Comparison to EQA benchmarks.}
\vspace{-10pt}\renewcommand{\arraystretch}{0.92}
\setlength{\tabcolsep}{5pt}
\begin{tabular}{l|cccccccc}
\hline
              & Simulator   & Dataset      & Real Scenes & Exploration Track & Track Numbers & Target Point & Question Creation & Open Vocab \\ \hline
EQA-v1 \cite{das2018embodied}        & House3D     & SUNCG        & \ding{55}       & \checkmark & - & \checkmark            & Rule-Based        & \ding{55}          \\
MP3D-EQA \cite{wijmans2019embodied}      & MINOS       & MP3D         & \checkmark           & \checkmark  &  -
 & \checkmark            & Rule-Based        & \ding{55}          \\
MT-EQA \cite{yu2019multi}    & House3D     & SUNCG        & \ding{55}    & \checkmark & - & \checkmark       & Rule-Based        & \ding{55}    \\
IQA \cite{gordon2018iqa}      & AI2THOR     & -       & \ding{55}    & \ding{55}  & - & \ding{55}            & Rule-Based        & \ding{55}       \\
VideoNavQA \cite{cangea2019videonavqa}    & House3D     & SUNCG        & \ding{55}           & \ding{55}  &  -
 &  \ding{55}            & Rule-Based        & \ding{55}          \\
K-EQA \cite{tan2023knowledge}         & AI2Thor     & -            & \ding{55}           & \ding{55}  & - & \ding{55}            & Rule-Based        & \ding{55}          \\
HM-EQA \cite{ren2024explore}        & Habitat     & HM3D         & \checkmark           & \ding{55}  & - & \ding{55}            & VLMs              & \ding{55}        \\
S-EQA \cite{dorbala2024s}         & VirtualHome & -            & \ding{55}           & \ding{55}   & - & \ding{55}            & LLMs              & \ding{55}            \\
NoisyEQA \cite{wu2024noisyeqa}      & -           & -            & -           & \ding{55}   & -  & \ding{55}            & VLMs              & \checkmark            \\
CityEQA \cite{zhao2025cityeqa}  &  EmbodiedCity  &  -  & \ding{55}   &  \ding{55}  &  - &   \checkmark    &    Manual   &   \checkmark  \\
\rowcolor{red!10}OpenEQA \cite{majumdar2024openeqa}       & Habitat     & ScanNet/HM3D & \checkmark           & \checkmark  &  152
 &  \ding{55}            & Manual            & \checkmark         \\
\rowcolor{green!10}\textbf{EXPRESS-Bench (Ours)} & Habitat     & HM3D         & \checkmark           & \checkmark  &  777 & \checkmark            & VLMs              & \checkmark            \\ \hline
\end{tabular}
\vspace{-15pt}
\label{Table1}
\end{table*}

\section{Introduction}
Embodied Question Answering (EQA) is a pivotal challenge at the intersection of computer vision, natural language processing, and embodied intelligence. In this task, an embodied agent must navigate a 3D environment, actively gather visual information through exploration, and answer questions about the scene~\cite{das2018embodied, majumdar2024openeqa}. Unlike traditional question-answering (QA) systems that rely on static images~\cite{antol2015vqa,ishmam2024image} or pre-existing knowledge bases~\cite{allam2012question, kwiatkowski2019natural}, EQA demands sequential decision-making, where agents must dynamically explore their surroundings to acquire the necessary information before formulating an answer. 
Thus, traditional QA methods fail to generalize to EQA because they are not designed to handle the dynamic, multi-step reasoning and embodied navigation essential for EQA \cite{zheng2024towards,song2024towards}.
This unique combination of perception, reasoning, and action makes EQA a compelling yet challenging problem with implications for real-world applications such as robotics, virtual assistants, and autonomous navigation~\cite{liu2024aligning}.

Despite its potential, current EQA methods face a fundamental limitation: models often generate answers without truly grounding them in exploration~\cite{majumdar2024openeqa}. This issue stems from the reliance on manually created  \cite{das2018embodied} or rule-based datasets \cite{wijmans2019embodied}, which are costly, inflexible, and prone to bias. As a result, agents frequently exploit spurious correlations rather than engaging in meaningful reasoning, raising concerns about their credibility in real-world applications \cite{luo2022depth,liu2023cross}. Additionally, ambiguity in question design and scenario complexity often lead to non-unique answers, making performance evaluation unreliable~\cite{luo2025dspnet}. More critically, some datasets embed prior knowledge, enabling models to generate answers without actual exploration—an issue known as unfaithful question answering~\cite{zhang2024tree}. For example, explicitly mentioning a task location (e.g., “living room”) allows models to guess answers without interacting with the environment. Evaluation metrics further compound these challenges. Existing metrics fail to assess answer reliability and do not effectively detect model hallucinations—where responses seem plausible but are incorrect \cite{chen2025cross,jiang2024hal}. This leads to biased assessments that overlook an agent's true exploration and reasoning abilities. Moreover, commonly used metrics, which primarily measure answer similarity, lack the granularity needed to evaluate exploration efficiency and quality, limiting their applicability to real-world scenarios~\cite{leng2024mitigating}. These gaps highlight the urgent need for more comprehensive EQA benchmarks that assess both the exploration process and answer quality.

In addition to the limitations of benchmark datasets and evaluation metrics, existing EQA methods also suffer from a critical deficiency: inefficient exploration capabilities.
Most current EQA models rely on frontier-based exploration~\cite{yamauchi1997frontier,yamauchi1998frontier,freda2005frontier,holz2010evaluating}, where agents expand their search by identifying and navigating towards unexplored frontiers on a map. While effective in open spaces, this approach struggles in constrained environments such as narrow corridors or cluttered scenes, limiting the agent’s capacity for comprehensive exploration~\cite{gupta2013interactive,niroui2019deep}. Furthermore, frontier-based strategies often lead to inefficient behaviors, such as redundant revisits to semantically important regions without acquiring new information. Crucially, existing methods lack mechanisms for fine-grained exploration of task-relevant areas, preventing agents from gathering the detailed environmental context needed for accurate reasoning~\cite{wang2024improved}. Therefore, it is essential to integrate exploration efficiency with answer reliability to ensure meaningful reasoning.

To address these challenges, we introduce the exploration-aware EQA task, which emphasizes the need for agents to actively and rationally explore relevant environmental clues before answering. To comprehensively assess embodied agents' exploration abilities, we construct EXPRESS-Bench, a large-scale benchmark comprising 777 exploration trajectories and 2,044 question-trajectory pairs—offering superior coverage compared to OpenEQA (as shown in Tab.\ref{Table1}). To enhance exploration efficiency, we propose Fine-EQA, a hybrid method that integrates frontier-based and goal-oriented exploration. By leveraging a global semantic map and functional region semantic map, Fine-EQA enables agents to efficiently navigate entire scenes while conducting fine-grained investigations of task-relevant areas. To ensure faithful answer evaluation,  we introduce the Exploration-Answer Consistency (EAC) metric, which explicitly measures the alignment between an agent’s exploration process and its generated answers. EAC effectively detects ungrounded responses that appear correct but lack supporting exploration evidence, providing a more rigorous assessment of model performance. Our contributions are summarized as follows:

\begin{figure*}[t]
	\centering
    \includegraphics[width=1\textwidth]{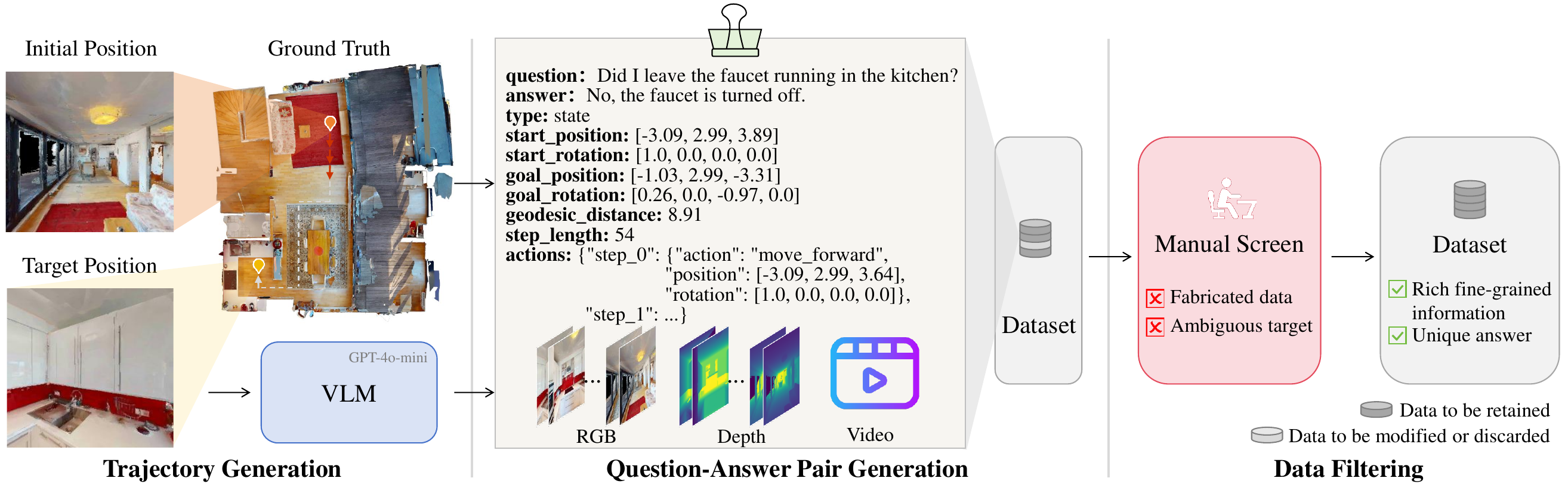}
         \vspace{-20pt}
	\caption{The construction process of EXPRESS-Bench.}
     \vspace{-15pt}
	\label{fig:frame}
\end{figure*}

\begin{itemize}
    \item   We introduce EXPRESS-Bench, a high-quality, large-scale EQA dataset designed to support active exploration while addressing key limitations of existing datasets, such as ambiguity and over-reliance on prior knowledge.
    \item We propose Fine-EQA, a two-stage exploration framework that enhances navigation efficiency and enables fine-grained exploration of task-relevant regions with flexible strategy switch, setting a new baseline for EQA.
    \item To ensure a faithful assessment of EQA, we introduce the EAC metric, which simultaneously evaluates answer grounding and exploration efficiency, providing a more rigorous measure of the model performance.
\end{itemize}

\section{Related Work}
\subsection{EQA Dataset}

Compared to visual question answering (VQA) datasets, EQA datasets are constructed in a three-dimensional space, encompassing both static scene descriptions and dynamic interactions, making their development a significantly more challenging task. 
Early works~\cite{das2018embodied, wijmans2019embodied, yu2019multi, gordon2018iqa, cangea2019videonavqa, tan2023knowledge} adopted a template-based approach to accelerate data generation; however, this method often led to simplistic question formats and straightforward answers. With the advent of large models, \cite{ren2024explore,dorbala2024s,wu2024noisyeqa} leveraged their capabilities to facilitate a more efficient dataset construction process that better aligns with research needs, thereby greatly enhancing data diversity and richness. \cite{majumdar2024openeqa} proposed an open-ended dataset through manual design, demonstrating innovation but primarily focusing on scenario-based memory questions while overlooking the critical role of active exploration in EQA tasks. \cite{zhao2025cityeqa} extended the task to city spaces, incorporating the complexities of urban environments. 
However, the complexity of 3D environments often leads to non-unique answers, making model evaluation challenging. Besides, existing datasets rarely support active exploration. Therefore, we propose an exploration-aware EQA benchmark that enables active exploration and ensures unique answers, providing high-quality evaluation and numerous EQA tasks.

\subsection{Large Models for Embodied Agents}
The strong reasoning and generalization capabilities of large models have driven their widespread adoption in embodied tasks, such as vision-language navigation \cite{long2024instructnav,zheng2024towards,long2024discuss} and embodied manipulation \cite{wu2023tidybot,kapelyukh2024dream2real,xu2024rt}. However, EQA demands more comprehensive reasoning, encompassing not only navigation and interaction but also question answering based on environmental information. This heightened requirement for semantic understanding makes it difficult to directly transfer methods from other tasks.
In EQA, some studies utilize these models for target object detection \cite{sakamoto2024map} and subtask planning \cite{zhao2025cityeqa}. They are also employed to assign confidence scores to exploration directions \cite{ren2024explore,cheng2024efficienteqa} or generate semantic labels \cite{saxena2024grapheqa} to update semantic maps, integrating a frontier-based exploration strategy. However, scene complexity often constrains an agent's exploration, resulting in inefficiencies or hallucinated responses due to missing critical information. To address this, we propose a two-stage EQA model that enables a flexible strategy switch between frontier-based and goal-oriented exploration, enhancing efficiency and robustness.

\section{EXPRESS-Bench}
In this section, we describe how to construct the EXPRESS-Bench utilizing the multimodal large model GPT-4o-mini. Unlike existing EQA datasets, we not only record the generated questions and ideal responses, but also record the ground truth actions, as well as the state and information acquired by the agent after each operational step.

\subsection{Simulator}
HM3D is a comprehensive dataset featuring 3D reconstructions of 1,000 large-scale buildings from a wide range of real-world locations. We utilize the portion of the HM3D dataset that are rich in semantic information on the Habitat simulator to replicate real-world environments, and make extensive use of the Habitat interface to construct the data.

\subsection{Dataset Generation Pipeline}

In this section, we provide an overview of the question-answer pairs creation pipeline to create EXPRESS-Bench. The dataset construction process is illustrated in Fig.\ref{fig:frame}. 

\textbf{Stage 1: Trajectory Generation.}
We randomly sample navigable initial and target positions within the scene. Then, we compute the shortest sequence of atomic actions (“move forward,” “turn left,” and “turn right”) and the corresponding number of steps required to reach the goal from the starting position, which serve as the ground truth. We also record the shortest geodesic distance between these positions. Given the large scale of the simulated environment, we constrain both the initial and target positions to the same floor and ensure the step length falls between 10 and 100. After each atomic action, the agent's state including coordinates, orientation and first-person visual observation is recorded. Based on the collected visual data, we further generate a trajectory video that visualizes the agent’s complete exploration process from a first-person perspective.

\textbf{Stage 2: Question-Answer Pair Generation.}
In EQA, the agent’s final visual observation contains crucial information needed to answer the question. Therefore, we feed the visual observation at the target location into the multimodal large model, along with several example question-answer pairs. These questions are designed from a human perspective to simulate natural conversations in everyday home scenarios.
Guided by the prompt, GPT-4o-mini generates both questions and corresponding answers. The answers are open-ended rather than simple yes/no or multiple-choice responses, which helps reduce data bias and prevents models from relying solely on their inherent common sense, placing higher demands on models’ performance. 

\textbf{Stage 3: Data Filtering.}
Although large models perform well, their outputs are not always reliable and reasonable. Additionally, due to the vast environment, there may be multiple rooms of the same type (e.g., bedrooms) or duplicate objects in the scene, which can result in non-unique answers to the questions. Therefore, manual screening of the generated data is necessary. 
First, we ensure that the answers to all questions remain relevant to the scene. 
Second, we utilize the generated trajectory video and the top-down views of each floor provided by the Habitat simulator to track the agent's position and movement in the scene. This allows us to address issues where the target region cannot be clearly identified or reached accurately during exploration. Specifically, we retain only those questions where either no other regions of the same type as the target region, or where the target region is the closest to the initial position among similar regions, ensuring the uniqueness of the answer. 
When necessary, we also specify key regions of the questions and include contextual details, such as the location, attributes, and relationships of the target object.
We invited several individuals to review the data. To ensure the dataset's high quality, a curator performed a final check to verify its overall consistency.

\subsection{Dataset Statistics}
\begin{figure}[t]
\centering
\includegraphics[width=0.48\textwidth]{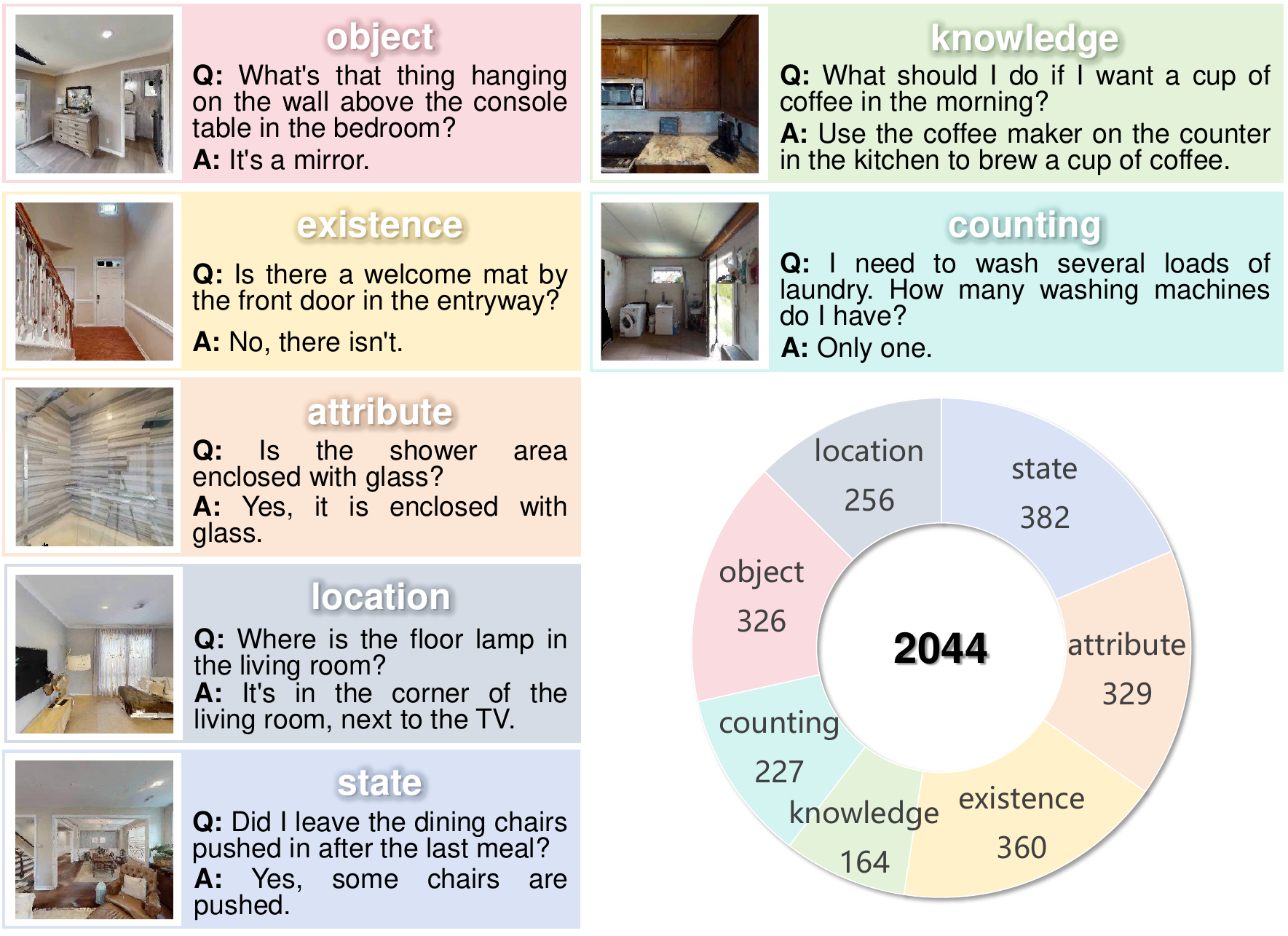}
\vspace{-20pt}
\caption{Overview of the EXPRESS-Bench statistics.}
 \vspace{-15pt}
\label{fig:distribution}
\end{figure}

Finally, the EXPRESS-Bench contains 777 trajectories, encompassing a total of 2,044 question-trajectory pairs. 
The EXPRESS-Bench predominantly consists of questions in the following seven categories: \textit{state}, \textit{knowledge}, \textit{location}, \textit{attribute}, \textit{counting}, \textit{existence}, and \textit{object}. The distribution of these categories is shown in Fig.\ref{fig:distribution}.
On average, the agent navigated the scene with a step size of 39.8 steps per question, covering a geodesic distance of 6.6 meters. More analysis of EXPRESS-Bench is in the Supplementary Material.

\subsection{Exploration-Answer Consistency Metric} 
\label{sec:metrics}
Since the answers are open-vocabulary, directly assessing their correctness is not feasible. OpenEQA leverages GPT-4 to evaluate the model's output, assigning a score from 1 to 5. However, OpenEQA overlooks the grounding of the answers—meaning the model's response may be fabricated and unrelated to the environmental information observed by the agent. To address this limitation, we refine the correctness evaluation score, ${\sigma}_i$, which assesses the model's response based on the question, the correct answer and the visual observation. Additionally, we introduce a grounding evaluation score, ${\delta_i}$, to assess the relevance of the answer to the environment with the support of the VLM:
\begin{equation}
\sigma_{i}=\varphi (Q_{i}, A_{i}^{*}, A_{i}, I_{i})
\end{equation}
\begin{equation}
    \delta_{i}=\psi ( Q_{i}, A_{i}, I_{i})
\end{equation}
where $Q_i$ is the given question, $A_i^*$ is the correct answer, $A_i$ is the model’s response, and $I_i$ is the environment image from the agent's final first-person observation. The score of an answer is calculated as $\sigma_{i} * \delta_{i}$.

Due to the open-ended nature of answers, when the model's response aligns with the question and image but differs from the correct answer, incorporating the observed image into the evaluation of ${\sigma}_i$ also enables the VLM to assign more precise scores.
Guiding prompt for the model's scoring process is provided in the Supplementary Material.

\begin{figure}[t]
\centering
\includegraphics[width=0.38\textwidth]{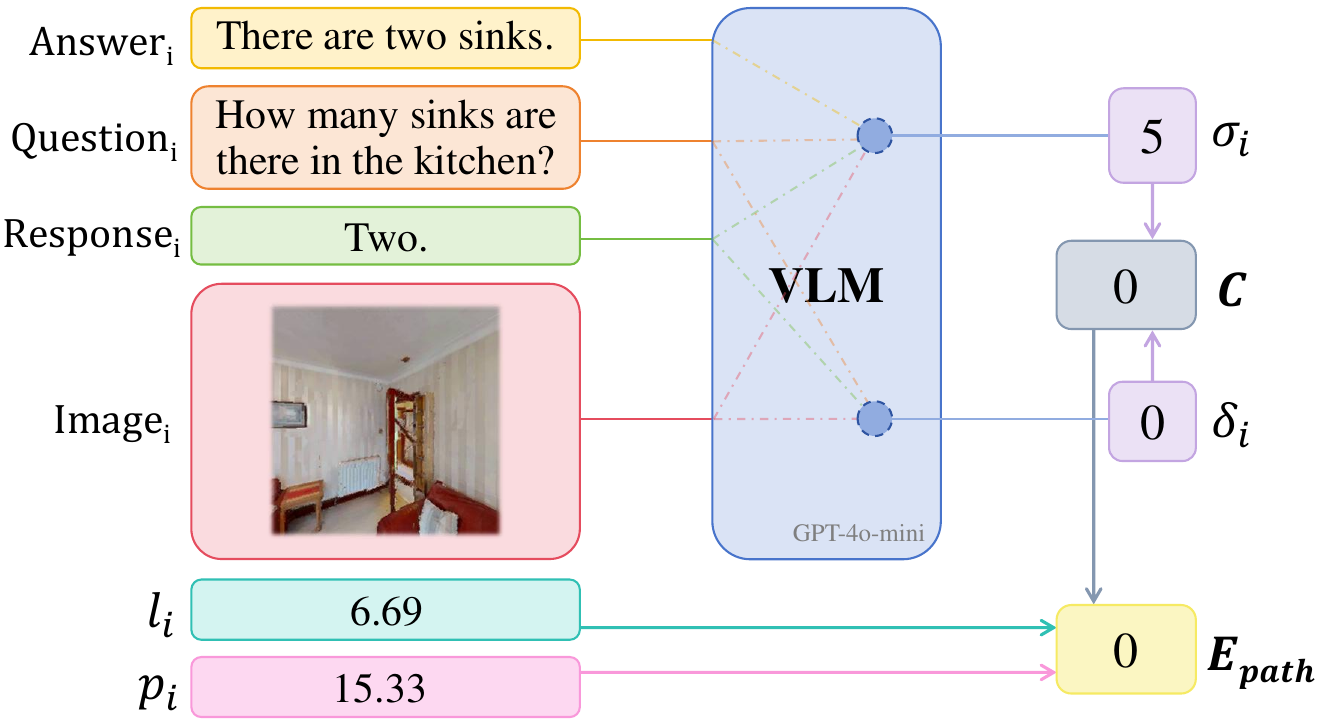}
 \vspace{-10pt}
\caption{Exploration-Answer Consistency Metric.}
 \vspace{-15pt}
\label{fig:evaluation}
\end{figure}

The VLM assigns the $\sigma_i$ score on a scale of 1 to 5, with higher scores indicating that the model's response is more accurate and closer to the correct answer. 
The $\delta_i$ score is assigned as 0, 0.5, or 1. A score of 1 signifies that the agent’s observation is relevant to the question and the model’s description of the environment is accurate, aligning with the agent’s perception. A score of 0.5 denotes relevant observation but an incorrect model description. A score of 0 implies irrelevant observation, with the model fabricating output unrelated to the environment, even if it matches the correct answer.
The evaluation pipeline is shown in Fig.\ref{fig:evaluation}. 

Next, we calculate the overall correctness $C$ of the output as well as the model's efficiency $E_{path}$ in completing the task:

\begin{equation}
    C= \frac{1}{N}\sum\limits_{i=1}^{N}\frac{{\sigma}_{i} \times {\delta}_{i}}{5}\times 100\%
    \end{equation}
\begin{equation}
    E_{path}= \frac{1}{N}\sum\limits_{i=1}^{N}\frac{{\sigma}_{i} \times {\delta}_{i}}{5}\times \frac{l_i}{max(p_i,l_i)}\times 100\%
\end{equation}
where $N$ is the total number of questions, $l_i$ represents the distance the agent navigate along the ground truth path that is sufficient to complete the task, and $p_i$ is the actual distance the agent moves during the experiment.

Furthermore, we assess the agent's navigation performance by measuring 
the geodesic distance $d_T$ from its final exploration position $P_E^i$ to the target location $P_T^i$: 
\begin{equation}
    d_T = \frac{1}{N}  {\sum \limits_{i=1}^{N}} dis_g(P_E^i, P_T^i) 
    \end{equation}
While the EQA task does not require the agent to reach the target as closely as navigation tasks, using this metric allows for a more thorough evaluation of the model.

\section{Fine-EQA}

\begin{figure*}[t]
	\centering
\includegraphics[width=0.85\textwidth]{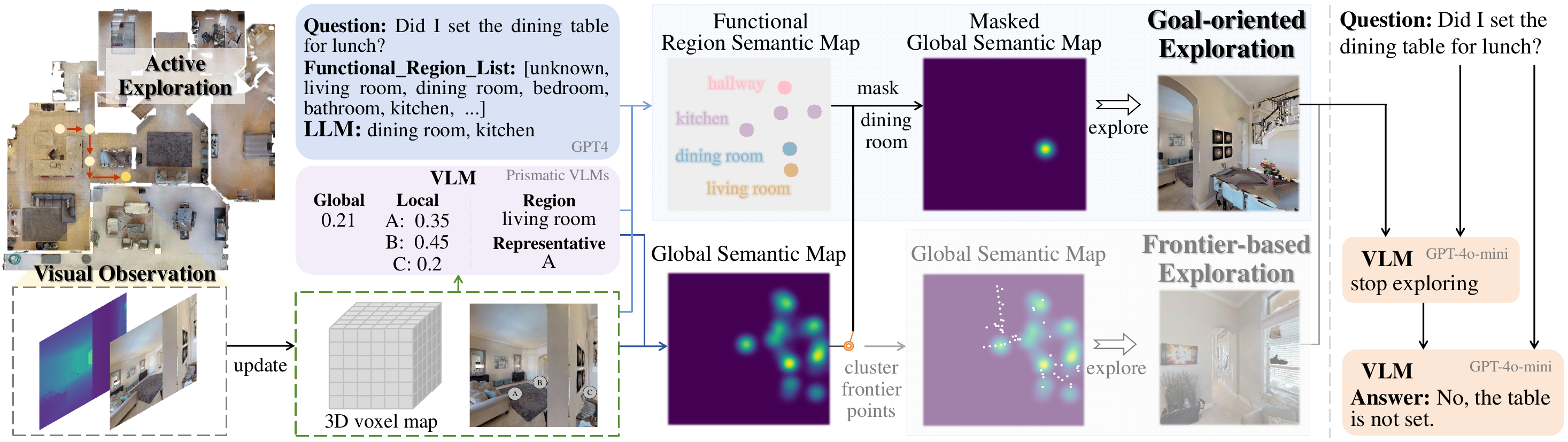}   \vspace{-10pt}
	\caption{The Fine-EQA framework operates as follows: The agent initially performs coarse-grained exploration using a frontier-based strategy, then switches to goal-oriented fine-grained exploration once task-relevant regions are identified. A maximum exploration limit per region prevents excessive searching, prompting the agent to either return to frontier-based exploration or focus on the next most promising region. Throughout this process, the VLM continuously evaluates the relevance and completeness of the acquired information, guiding the agent's decision to either continue exploration or generate answers based on the most recent visual inputs, as detailed in the Appendix.}
    \vspace{-15pt}
	\label{fig:model}
\end{figure*}

We propose Fine-EQA, a flexible two-stage EQA framework that strategically integrates frontier-based exploration with goal-oriented exploration. Fine-EQA constructs dual semantic representations, global semantic maps and functional region semantic maps, to guide agents through complex environments with greater efficiency while ensuring thorough investigation of task-relevant areas. 

\subsection{Overview}

The Fine-EQA operates in two distinct phases: exploration and question answering, as illustrated in Fig.\ref{fig:model}. Firstly, the agent builds and maintains a comprehensive environmental representation using a 3D voxel map ($M_3 \in R^{d_x \times d_y \times d_z}$), which is projected onto a 2D planar map ($M_2 \in R^{d_x \times d_y}$) to track exploration states and spatial occupancy. Here, $d_x$, $d_y$ and $d_z$ are initialized according to the dimensions of the scene in 3D space. During the exploration phase, the agent alternates between frontier-based and goal-oriented exploration, with a region-specific limit preventing excessive searching, while GPT-4o-mini decides when to stop and generate responses:

\begin{itemize}
    \item \textbf{Frontier-Based Exploration (FBE)} identifies boundaries between known and unknown regions to expand environmental understanding.
    \item \textbf{Goal-Oriented Exploration (GOE)} targets regions with high task relevance based on semantic analysis.
\end{itemize}



\subsection{Frontier-Based Exploration}
FBE is fundamental to systematic environmental discovery, focusing on the boundaries between explored and unexplored regions. 
Inspired by \cite{ren2024explore}, we enhance traditional frontier-based approaches by incorporating semantic understanding into the exploration process.
We construct a global semantic map $M_{sem} \in R^{d_x \times d_y}$ by integrating both global semantic values $v_g$ and local semantic values $v_l$. At each time step, the agent projects its current RGB observation onto the 2D map $M_2$ and applies farthest point sampling to identify navigable points that maximize spatial coverage. These sampled points $p_{sample}$ are back-projected onto the original RGB image, where our VLM evaluates their exploration priority based on task relevance, establishing $v_l$. 
Concurrently, we extract the exploration decision confidence score of the VLM for the scene depicted in the RGB image, denoted as $v_g$.
We compute the semantic value of $p_{sample}$ through the weighted fusion of $v_l$ and $v_g$ and dynamically update $M_{sem}$ using Gaussian smoothing:
\begin{equation}
    M_{sem} \gets  (p_{sample}, v_l, v_g)
\end{equation}

To identify frontier points, we analyze the exploration states surrounding each point in $M_2$ and perform clustering to obtain candidate frontier points $F = \{f_1, f_2, ..., f_n\}$. Each candidate frontier point $f_i$ receives a composite weight $w_i$ determined by:
\begin{equation}
    w_i = \omega (v_{sem}^i, r_e^i, r_o^i, dis(f_i, p_{cur}))
\end{equation}
where $v_{sem}^i$ represents the semantic value, $r_e^i$ and $r_o^i$ are the unexplored rate and the unoccupied rate along the exploration direction, respectively. $dis(f_i, p_{cur})$ is the Euclidean distance to the agent's current position $p_{cur}$. $\omega$ incorporates an exponentially weighted enhancement of $r_e^i$, $r_o^i$, and $v_{sem}^i$ near the frontier point, while simultaneously applying an exponential decay based on $dis(f_i, p_{cur})$. This encourages the agent to prioritize unexplored areas and minimize local redundancy, ultimately enhancing exploration efficiency.

The agent then selects the next location $\chi$ for exploration by treating these weights as probabilities:
\begin{equation}
    \chi = \gamma(F, W),~ 
    W=\{w_i / \sum_{j=1}^{N} w_j, i = 1, 2, ..., N\}
\end{equation}
where $\gamma$ represents the random sampling of points with different probabilities.

\subsection{Goal-Oriented Exploration}
The primary limitation of pure frontier-based exploration is its inability to comprehensively explore task-relevant regions,
particularly in complex environments with occlusions and spatial constraints. Our goal-oriented exploration strategy addresses this limitation by explicitly modeling functional regions and their relevance to the given question.

\noindent \textbf{Functional Region Semantic Mapping.} We construct a functional region semantic map $M_{reg}\in R^{d_x \times d_y}$ to guide the agent toward areas with high task relevance. During observation, we instruct the VLM to classify the current scene into specific functional regions (e.g., kitchen, dining room, bedroom) and identify representative points $q$ within these regions from the sampled points $p_{sample}$. Formally, when the VLM identifies a functional region $Reg$ with confidence exceeding a predefined threshold, we update $M_{reg}$ around the representative point:
\begin{equation}
    M_{reg}(N(q)) = \textrm{ID}_{Reg}
\end{equation}
where $N(q)$ is the neighborhood of point $q$ and $\textrm{ID}_{Reg}$ represents the semantic value associated with $Reg$. Points with identical semantic values in $M_{reg}$ that exhibit spatial proximity are merged into coherent regions, creating a comprehensive functional decomposition of the environment.

\noindent \textbf{Task-Relevant Region Prioritization.} We use an LLM to analyze the question and prioritize relevant regions based on their task importance. This prioritization is essential because: 1) secondary regions often provide important context or pathways to primary regions, and 2) exploring multiple regions strategically improves spatial awareness and navigation efficiency. It also enables explicitly modeling of scenarios with multiple task-related regions. When the agent identifies a higher-priority region, it shifts to goal-oriented exploration within that region.

\noindent \textbf{Masked Semantic Mapping.} To focus on high-priority regions, we apply masking operations $\phi$ to the global semantic map based on the functional region semantic map:
\begin{equation}
    M_{masked} = \phi (M_{sem}, M_{reg}, r)
\end{equation}
where $r$ is the region with the current highest priority. This masked semantic map $M_{masked}\in R^{d_x \times d_y}$ exclusively contains semantic values for the prioritized region, effectively directing the agent's attention toward task-relevant areas. To prevent redundant exploration, previously visited points receive decreased semantic values within $M_{masked}$. 
With the masked semantic map, the agent selects the position 
$\chi = \textrm{argmax}_{x, y}( M_{masked} )$ 
with the highest semantic value for further exploration. This adaptive switching between exploration strategies ensures efficient coverage of the environment and thorough investigation of task-relevant areas.

\section{Experiments}

We evaluate various models on our EXPRESS-Bench, reporting the accuracy of their responses. We also present the efficiency and navigation performance of agents equipped with exploration capabilities.

\subsection{Baselines}
We compare various models in a zero-shot setting, including 1) Blind LLMs, 2) Socratic Models, 3) Multi-Frame VLMs, 4) Exploring Agents, and 5) Human Performance.

\textbf{Blind LLMs.} 
The Blind LLMs refers to LLMs that produce answers based purely on questions, completely disregarding any contextual information from embodied scenarios, formulated as $A = \textrm{LLMs}(Q)$. Specifically, we employ DeepSeek-V3 \cite{liu2024deepseek,guo2025deepseek}, GPT-4 \cite{achiam2023gpt}, and LLaMA-3-8b \cite{dubey2024llama}.

\textbf{Socratic Models.} 
Using ground truth data $G$, we simulate an agent's navigation and exploration to obtain first-person visual observations, from which we randomly sample $k$ frames. Each frame $G_i$ is described by an image captioning model as $d_i = \textrm{Captioner}(Q, G_i)$, which are then concatenated into $D = [d_1, d_2, ..., d_k]$. We set $k = \lceil \frac{L}{10} \rceil$, where $L$ denotes the step length of the ground truth. The LLMs then generate answers based on the question and the concatenated descriptions, i.e., $A = \textrm{LLMs}(Q, D)$. For caption generation, we employ GPT-4o-mini\cite{hurst2024gpt} and LLaVA-v1.5-7b, and for the LLMs, we use DeepSeek-V3, GPT-4 and LLaMA-3-8b.

\textbf{Multi-Frame VLMs.} We randomly select k frames from the ground truth, which are subsequently fed into the VLMs, i.e., $A=\textrm{VLMs}(Q, G_1, G_2, ..., G_k)$. We employ GPT-4o-mini and LLaVA-v1.5-7b.

\textbf{Exploring Agents.} 
We use various exploration strategies $S$ to guide the agent, i.e. $\chi = \pi (S)$. The final answer is generated based on the visual information $I$ at the termination point, given by $A=\textrm{VLMs}(Q, I)$.

\textbf{(1) Random Exploration (RE). }
The agent randomly explores the scene by selecting a movement distance and direction, navigating to the nearest point. To avoid excessive exploration, GPT-4o-mini decides when to stop.

\textbf{(2) Frontier-Based Exploration (FBE). } We use a pure frontier exploration method that ignores semantic information, with GPT-4o-mini determining when to stop.

\textbf{(3) Goal-Oriented Exploration (GOE). }The agent starts with random exploration and switches to goal-oriented exploration upon reaching a task-relevant region. Similar to Fine-EQA, once the maximum exploration limit for a given region is reached, the agent transitions to either random exploration or explores a suboptimal region.

\textbf{(4) Fine-EQA (Ours).} For LLMs, we use GPT-4 to identify task-related functional regions. For VLMs, we employ Prismatic VLMs (Prism-DINOSigLIP 7B) \cite{karamcheti2024prismatic} to assign semantic values and evaluate functional regions. GPT-4o-mini is used to decide when to stop exploration and generate responses.

\textbf{Human Agent.} Five participants review the ground truth and provide answers, denoted as $A = \textrm{Human}(Q, G)$. Their scores are then averaged.

\begin{table}[t]
\caption{Performance comparison on EXPRESS-Bench.}
\vspace{-10pt}\scriptsize
\label{table:performance}
\centering
\setlength{\tabcolsep}{7pt}
\begin{tabular}{lcccc}
\hline
                & $C$↑ & $C^*$↑    & $E_{path}$↑ &  $d_T$↓    \\ \hline
\rowcolor{gray!30} Human Agent           &   -     & 83.99     & -        & -       \\ \hline
\multicolumn{5}{l}{\textbf{Blind LLMs}}                                         \\
DeepSeek-V3     & -    & 59.15        & -       & -   \\
GPT4           & -  & 58.96  & -         & -       \\
LLaMA-3-8b     & -  & 57.25  & -           & -       \\ \hline
\multicolumn{5}{l}{\textbf{Socratic Models}}                                    \\
DeepSeek-V3 w/ GPT-4o-mini             &    -    &   62.60
    &    -      &   -     \\ 
GPT4 w/ GPT-4o-mini                &    -    &    62.56
   &       -   &   -     \\ 
LLaMA-3-8b  w/ GPT-4o-mini              &   -     &    59.95
    &    -      &  -    \\
DeepSeek-V3 w/ LLaVA-v1.5-7b             &    -    &   60.63
    &    -      &   -     \\ 
GPT4 w/ LLaVA-v1.5-7b                &    -    &    59.53
   &       -   &   -     \\ 
LLaMA-3-8b  w/ LLaVA-v1.5-7b              &   -     &   58.59
    &    -      &  -    \\
\hline
\multicolumn{5}{l}{\textbf{Multi-Frame VLMs}}                                   \\
GPT-4o-mini    & -  & 58.37     & -        & -       \\
LLaVA-v1.5-7b   & - & 57.66     & -        & -       \\ \hline
\multicolumn{5}{l}{\textbf{Exploring Agents}}                                    \\
RE              &   36.95     &   62.75      &    12.06      &  7.26      \\ 
FBE            & 38.60  & 60.24 &   14.55   & 6.64  \\
GOE     &  38.54  &  63.34  &   12.74    &  6.46  \\
\rowcolor{green!10}\textbf{Fine-EQA (Ours)}    &  \textbf{40.55}   &  \textbf{63.95}  &   \textbf{16.22}    &  \textbf{6.43}  \\ \hline
\end{tabular}
\vspace{-15pt}
\end{table}

\begin{figure*}[t]
\centering
\includegraphics[width=0.95\textwidth]{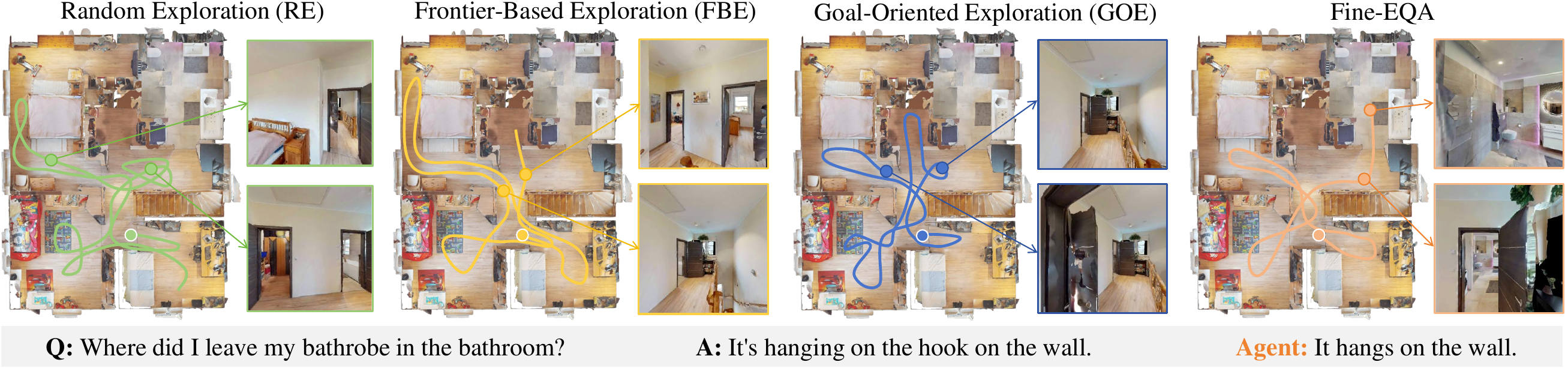}
\vspace{-10pt}
\caption{Exploration trajectories of agents employing different strategies within the scene. Only our Fine-EQA effectively guides the agent to observe question-relevant regions and generate the correct answer.}
\vspace{-10pt}
\label{fig:exploration}
\end{figure*}

\begin{figure*}[t]
\centering
\includegraphics[width=0.85\textwidth]{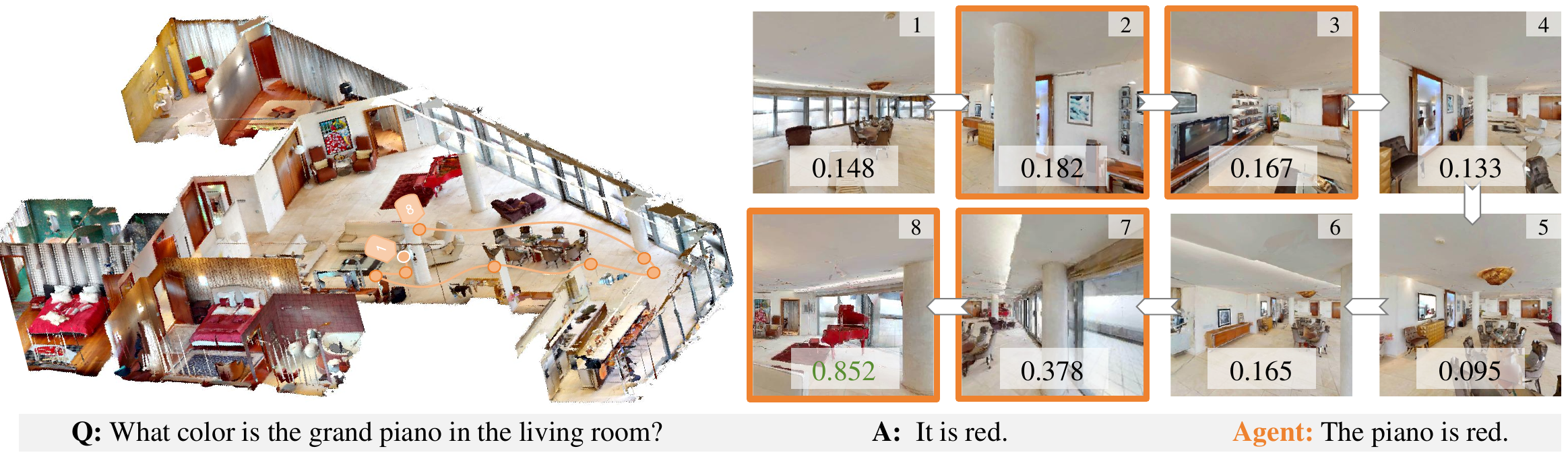}
\vspace{-10pt}
\caption{Confidence of visual observations at waypoints during the agent’s exploration. The highest confidence is indicated in green in the final frame, demonstrating the faithfulness of our Fine-EQA by maintaining exploration-answer consistency.}
\vspace{-10pt}
\label{fig:answer}
\end{figure*}

\subsection{Comparison with State-of-the-art Methods}

Besides the metrics in Section \ref{sec:metrics}, we also compute $C^*$, which is $C$ without answer grounding (i.e., setting $\delta_i = 1$). Since models like GPT-4 cannot perform exploration, we only assess their performance using $C^*$, as shown in Tab.\ref{table:performance}.

\noindent\textbf{Non-Exploratory Agents.} Among LLMs without visual information, DeepSeek-V3 demonstrates the best performance. When combined with VLMs for reasoning, their performance is improved, with DeepSeek-V3-based Socratic agents yielding the highest results. However, since EXPRESS-Bench focuses on active exploration, with questions typically related to the last frame, the performance improvement of the Socratic Models, which relies solely on passively observed video frames, remains limited.
Additionally, Multi-Frame VLMs performs worse than the Socratic models, indicating that critical information extraction is essential for exploration-aware EQA than simply using numerous images. This supports the reliability of EXPRESS-Bench in evaluating active exploration capabilities.

\noindent\textbf{Active Exploration Agents.} Agents with active exploration capabilities demonstrate enhanced environmental perception and outperform nearly all non-exploratory models. 
RE explores the environment without constraints until it either finds the answer or reaches the maximum exploration limit, leading to a high $C^*$ value. However, its $C$ value drops significantly, while its $d_T$ and $E_{path}$ metrics perform worse than other methods. This suggests that its actual performance is lower than what $C^*$ implies and is influenced by model hallucinations.  
In contrast, FBE and GOE leverage their respective advantages—FBE excels in exploring unknown environments, while GOE ensures thorough exploration of target regions. Consequently, they achieve superior performance in exploration efficiency ($E_{path}$) and navigation ($d_T$), with FBE slightly outperforming GOE in the $C$ metric. 
By strategically integrating frontier-based exploration with goal-oriented exploration, our Fine-EQA surpasses all baselines across these metrics, demonstrating the effectiveness of our exploration strategy. In terms of navigation performance, Fine-EQA performs well on $d_T$, indicating its ability to effectively approach target regions and enhance visual perception. While the EQA task does not require precise target positioning, the results suggest that agents closer to the target tend to provide more accurate answers, which aligns with intuition.


\noindent\textbf{Human Agent.}
Although Fine-EQA performs well across various metrics, all methods still exhibit a significant gap in $C^*$ scores compared to human performance. This result validates the effectiveness of EXPRESS-Bench and its evaluation metrics while highlighting the limitations of current methods in complex environments, emphasizing the need for further improvements of EQA methods.

\subsection{Experiments on Other Datasets}
To further evaluate the performance of Fine-EQA, we conduct experiments on two additional datasets. 
For the OpenEQA\cite{majumdar2024openeqa}, we focus specifically on the A-EQA subset, which assesses the agent's ability to explore the environment and answer questions. We use the corresponding evaluation metrics $C'$ and $E'$ for performance measurement, with their formulas provided in the Supplementary Material.

The results are presented in Tab.\ref{table:openeqa}. Fine-EQA outperforms the best-performing GPT-4V model from \cite{majumdar2024openeqa}, particularly in terms of exploration efficiency. This is because the active exploration strategy in \cite{majumdar2024openeqa} relies entirely on the frontier-based method and fails to terminate exploration promptly after gathering the information necessary for the task.

\begin{table}[]
\caption{Performance comparison on the A-EQA subset of OpenEQA. Results marked with * are from the OpenEQA benchmark, where GPT-4V is evaluated on a random subset of 184 questions. In contrast, our Fine-EQA is evaluated on the full set of questions.}
\vspace{-10pt}\scriptsize
\label{table:openeqa}
\centering
\begin{tabular}{lcc}
\hline
         & $C'$↑          & $E’$↑         \\  \hline
OpenEQA w/ GPT-4V   & 41.8$_{\pm3.2}$* & 7.5$_{\pm0.6}$* \\
Fine-EQA & \textbf{43.27}       & \textbf{29.16}    \\  \hline 
\end{tabular}
\end{table}

We also evaluated the performance of Explore-EQA\cite{ren2024explore} and Fine-EQA on the multiple-choice dataset HM-EQA\cite{ren2024explore}, using answer accuracy and the length of the agent's navigation path as metrics. As shown in Tab.\ref{table:hmeqa}, Fine-EQA achieves substantial improvements over Explore-EQA in both metrics.

\begin{table}[]
\caption{Performance comparison on HM-EQA.}
\label{table:hmeqa}
\vspace{-10pt}\scriptsize
\centering
\begin{tabular}{lcc}
\hline
            &Accuracy(\%)↑ & Path Length(m)↓ \\ \hline
Explore-EQA & 50.4   & 93.687  \\
Fine-EQA    & \textbf{56.0}   & \textbf{54.267}  \\ \hline
\end{tabular}
\end{table}

Furthermore, we analyzed the ambiguity rate in questions, finding 35.6\% in HM-EQA, 62.3\% in Open-EQA, and only 9.7\% in EXPRESS-Bench. The absence of complete exploration trajectories or clear target locations in HM-EQA and Open-EQA further hinders accurate performance evaluation. In contrast, our EXPRESS-Bench enhances evaluation reliability and interpretability.

\noindent\subsection{Ablation Studies}
As shown in Tab.\ref{table:ablation}, the ablation of both FBE and GOE leads to a decline in performance, affecting both response accuracy and exploration efficiency. Specifically, the FBE module has a more significant impact, highlighting its crucial role in enhancing information acquisition.

\begin{table}[t]
\caption{Ablation Study of Model Modules.}
\vspace{-10pt}\scriptsize
\label{table:ablation}
\centering
\setlength{\tabcolsep}{10pt}
\begin{tabular}{lcccc}
\hline
                & $C$↑ & $C^*$↑    & $E_{path}$↑ &  $d_T$↓   \\ \hline
Fine-EQA w/o FBE     &  38.54  &  63.34  &   12.74    &  6.46  \\
Fine-EQA w/o GOE     &  39.63  &  60.74  &    14.64   &  6.54  \\
Fine-EQA     &  \textbf{40.55}  &  \textbf{63.95}  &   \textbf{16.22}    &  \textbf{6.43}  \\ \hline
\end{tabular}
\vspace{-5pt}
\end{table}

To isolate the impact of GPT-4o-mini on Fine-EQA’s performance, we replaced it with other VLMs and conducted evaluations on the EXPRESS-Bench.
As shown in Tab.\ref{table:llms}, Fine-EQA built using different VLMs exhibit varying performance across different metrics, but consistently outperform other models. We observed that Fine-EQA models with higher $C$ scores tend to engage in more extensive exploration within the environment, which is reflected in their lower $E$ scores.

\begin{table}[]
\caption{Ablation Study of VLMs.}
\label{table:llms}
\centering
\vspace{-10pt}\scriptsize
\begin{tabular}{lcccc}
\hline
            & $C$↑ & $C^*$↑    & $E_{path}$↑ &  $d_T$↓    \\ \hline
Janus-Pro-7B     &   39.88    &    63.35   &    20.86   &   6.18   \\
Qwen-Vl-Plus     &   40.02    &    63.06   &      17.57   &   6.41   \\ 
 GPT-4o-mini & 40.55 & 63.95 & 16.22 & 6.43 \\ \hline
\end{tabular}
\end{table}

\subsection{Effectiveness of Exploration and Answering}
We perform both qualitative and quantitative studies to evaluate the effectiveness of our exploration strategy and the faithfulness of our question reasoning module.

\noindent\textbf{Exploration Effectiveness.} In Fig.\ref{fig:exploration}, we randomly selected a scene and visualized the agent's exploration trajectory under different exploration strategies. RE disregards both scene and instruction information, selecting the next location arbitrarily. This leads to significant redundancy and inefficiency. 
FBE and GOE demonstrate improved performance over RE. However, they still struggle to effectively explore the scene, failing to cover relevant regions or gather sufficient contextual information.
Fine-EQA further enhances exploration efficiency by integrating scene context and the regional cues embedded in the instructions. While rapidly expanding the unknown areas, it strategically directs the agent toward task-relevant regions for a more thorough and efficient exploration. As shown in Fig.\ref{fig:exploration}, our Fine-EQA agent, starting in the hallway, identified the relevant bathroom ahead. It then navigated strategically towards the bathroom, ultimately reaching the correct answer.

\noindent\textbf{Answer Faithfulness.} 
Moreover, we utilize VLMs to assess the faithfulness of our question reasoning module. Specifically, we use Prismatic VLMs to determine whether the environmental observations collected during the agent's exploration support an accurate response to the question. From Fig.\ref{fig:answer}, the observation image captured at the agent’s termination point has the highest confidence. We selected the four images with the highest confidence, highlighted with orange borders, as inputs for LLaVA and GPT-4o-mini. Both models agreed that the final image provides the most informative basis for answering the question.

\begin{table}[t]
\caption{Quantitative comparison of model performance.}\scriptsize\vspace{-10pt}
\label{table:confidence}
\centering
\setlength{\tabcolsep}{12pt}
\begin{tabular}{cccc}
\hline
         & $NPL$↑  & $ACE$↑  &   $WCE$↑    \\ \hline
RE       &  0.23    &      0.55  &  0.16 \\
FBE      &  0.28  &      0.58    &  0.19 \\
GOE       &  0.27   &     0.58   &  0.18 \\  
Fine-EQA  &   \textbf{0.34} &      \textbf{0.59}    &  \textbf{0.22} \\ \hline
\end{tabular}
\vspace{-15pt}
\end{table}

We also computed the average confidence ($ACE$), normalized path length ($NPL$), and the normalized path length-weighted confidence ($WCE$) of the agent's observations at the termination location under various exploration strategies. The detailed formulas of these metrics are provided in the Supplementary Material. According to the quantitative results in Tab.\ref{table:confidence}, Fine-EQA demonstrates the best performance, verifying the faithfulness of our Fine-EQA by concurrently improving the agent's exploration efficiency and maintaining exploration-answer consistency. 

\section{Conclusion}

In this paper, we introduce EXPRESS-Bench, the largest dataset for evaluating both exploration and reasoning in EQA. We also propose Fine-EQA, a hybrid exploration model that improves exploration efficiency by combining frontier-based and goal-oriented navigation. Additionally, we introduce the Exploration-Answer Consistency (EAC) metric to better assess exploration and reasoning alignment. Extensive experiments demonstrate the effectiveness of EXPRESS-Bench in advancing exploration-aware EQA.

\clearpage
{
    \small \bibliographystyle{ieeenat_fullname}
    \bibliography{main}
}

\clearpage
\setcounter{page}{1}
\maketitlesupplementary


\section{Question Reasoning Module}
\label{section:question_reasoning}

\textbf{State Judgment of Exploration.}
In EQA, the agent need to accumulate environmental information through dynamic interactions to achieve accurate responses. Crucially, this exploration process requires termination within reasonable constraints rather than continuing indefinitely. At each step, the agent performs a sufficiency evaluation of its acquired information to determine whether to end the exploration and proceed to the answer generating phase before reaching the predetermined maximum interaction threshold. Thus, we use the visual language model that systematically integrates real-time visual observations with textual query semantics to comprehensively analyze the relevance and information adequacy of the scene. Once the VLM determines that all essential information has been gathered and no further exploration is needed to answer the question, it signals the conclusion of the exploration phase. At this point, the exploration state is marked as ``completed" and the agent transits to the QA phase.

\textbf{Answer Generation.}
This process relies on the latest visual information obtained in the exploration phase and the understanding of the question. During the reasoning process, the VLM integrates the language information and the image features from the previous exploration to generate the answer that conforms to the question semantics and the actual situation of the scene.

\vspace{-3pt}
\section{Extra Experiments}
\label{subsection:extra_experiments}
\subsection{Experimental Setup}
\label{subsection:experimental_setup}
\begin{itemize}
    \item  Maximum exploration limit. The agent's total explorations within a scene are proportional to the scene size, while consecutive explorations within a task-relevant region are limited to three. Only in designated regions does the agent observe from four directions—front, back, left, and right—ensuring a comprehensive view.
    \item Maximum step length. The agent's next exploration point must be within 3 meters of its current location, ensuring controlled movement within the scene.
\end{itemize}

\subsection{Experimental Metrics}

\subsubsection{Formulas for Metric Calculation}
\label{subsection:formulas}
$C^*$ is the performance metric that ignores answer grounding(i.e., setting $\delta_i$ = 1):
\begin{equation}
    C^*= \frac{1}{N}\sum\limits_{i=1}^{N}\frac{{\sigma}_{i} }{5}\times 100\%
\end{equation}

The calculation formulas for metrics on the OpenEQA are as follows:

\begin{equation}
    C'= \frac{1}{N}\sum\limits_{i=1}^{N}\frac{{\sigma}_{i}^{'} -1 }{4}\times 100\%
    \end{equation}
\begin{equation}
    E'= \frac{1}{N}\sum\limits_{i=1}^{N}\frac{{\sigma}^{'}_{i} -1 }{4}\times \frac{l_i}{max(p_i,l_i)}\times 100\%
\end{equation}

where $\sigma_i^{'}$ is determined by the LLMs based on the OpenEQA prompt, $l_i$ represents the distance the agent navigate along the ground truth path that is sufficient to complete the task, and $p_i$ is the actual distance the agent moves during the experiment.

The calculation formulas for metrics of reliability study are as follows:
\begin{equation}
    ACE = \frac{1}{N}  { \sum \limits_{i=1}^{N}} ce_i
\end{equation}

\begin{equation}
    NPL = \frac{1}{N}  { \sum \limits_{i=1}^{N}} \frac{l_i}{max(p_i,l_i)}
\end{equation}

\begin{equation}
    WCE = \frac{1}{N}  { \sum \limits_{i=1}^{N}} ce_i \times \frac{l_i}{max(p_i,l_i)}
\end{equation}

where $ce_i$ represents the confidence of the VLM for the image.

\begin{figure*}[!t]
\centering
\includegraphics[width=0.9\textwidth]{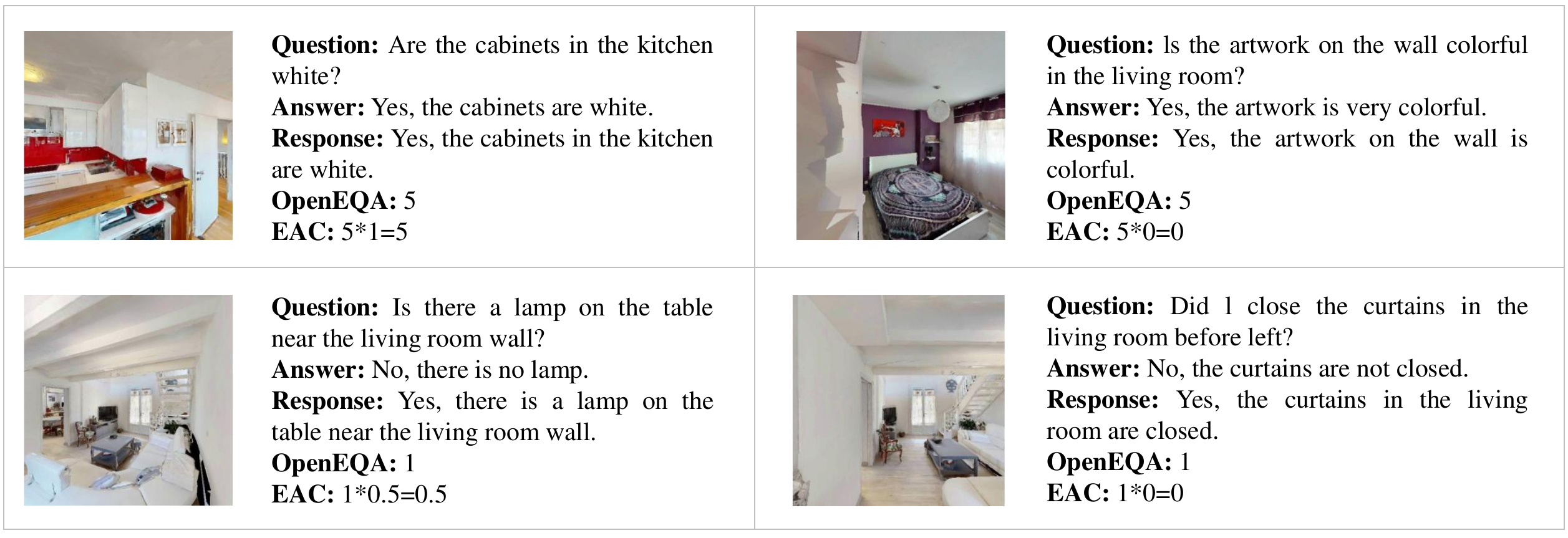}
\caption{Comparison of the metrics proposed by OpenEQA and ours. The EAC metric combines $\sigma$ and $\delta$ to jointly assess both the semantic validity and visual grounding of the response. By considering the grounding of the response, our metric offers a more reliable assessment of the model's performance.}
\label{fig:examples}
\end{figure*}

\begin{figure*}[t]
\centering
\includegraphics[width=0.65\textwidth]{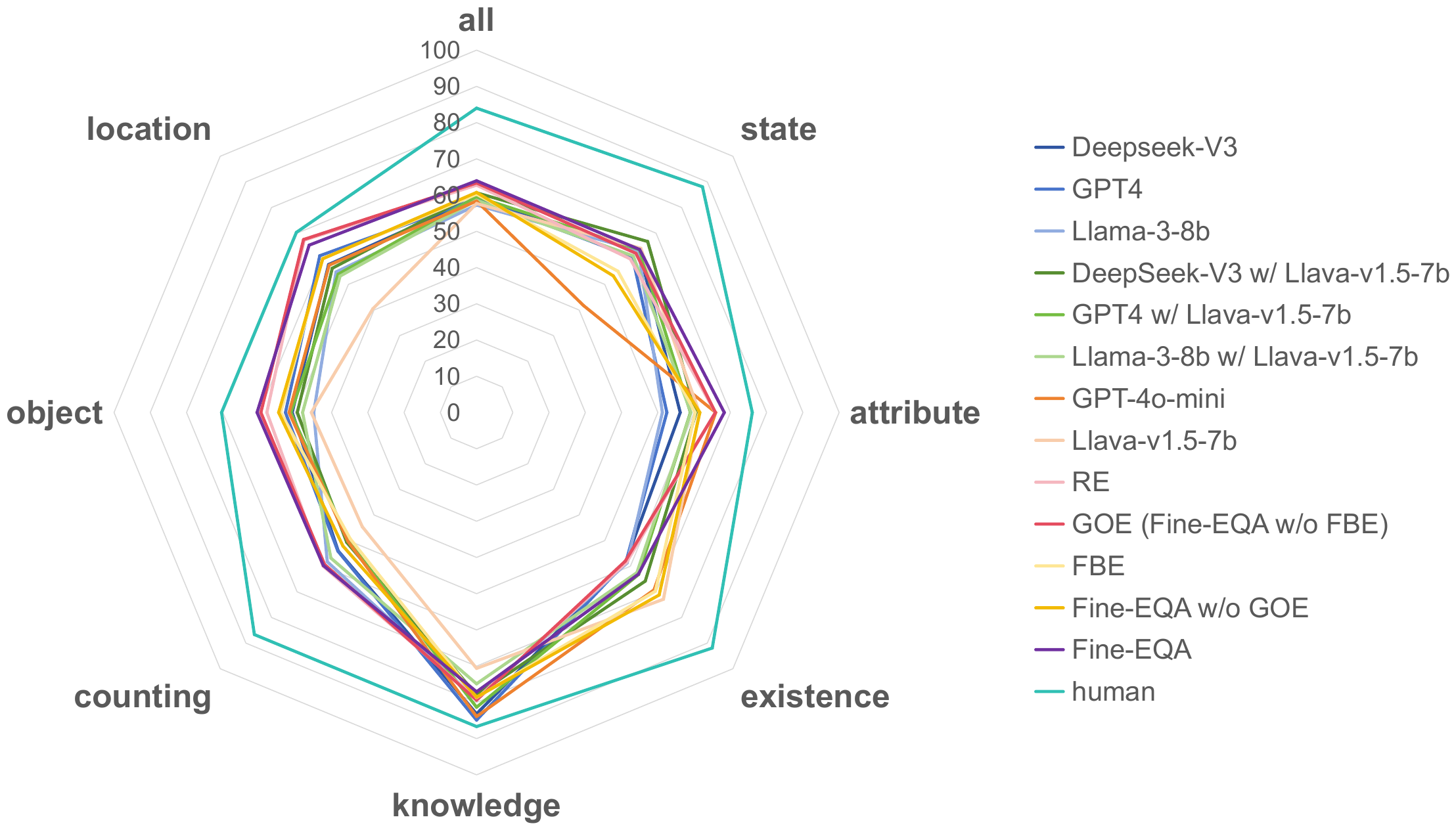}
\caption{Performance of models in the $C^*$ metric across different question types.}
\label{fig:performance1}
\end{figure*}

\begin{figure*}[!h]
\centering
\includegraphics[width=0.65\textwidth]{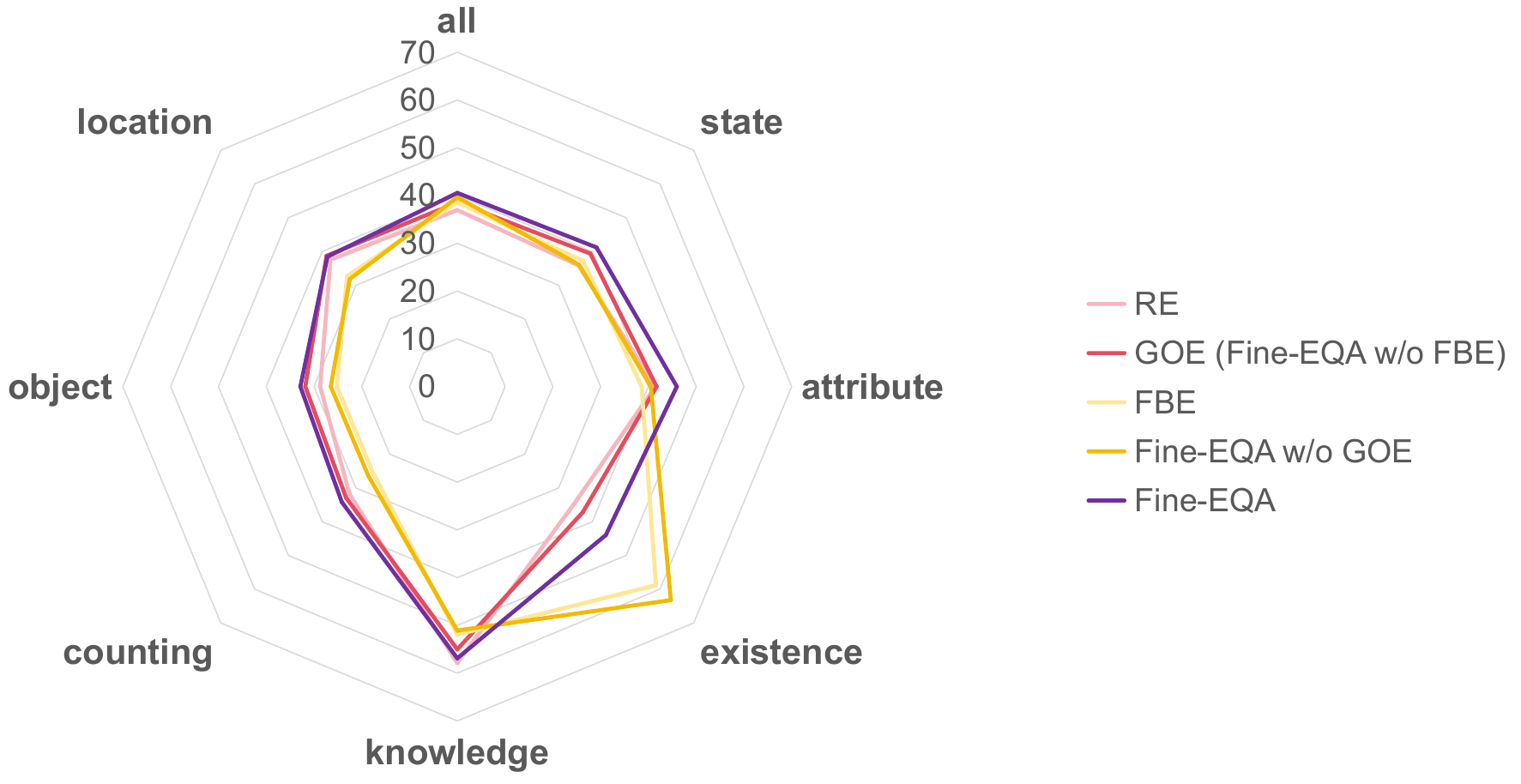}
\caption{Performance of exploration-aware agents in the $C$ metric across different question types.}
\label{fig:performance2}
\end{figure*}

\subsubsection{Comparison of Metrics}
\label{subsection:comparison}
By incorporating answer grounding, our metric provides a more accurate evaluation of the model's performance. Fig.\ref{fig:examples} compares our metric with that of OpenEQA using several examples.

\begin{figure*}[t]
\centering
\includegraphics[width=1\textwidth]{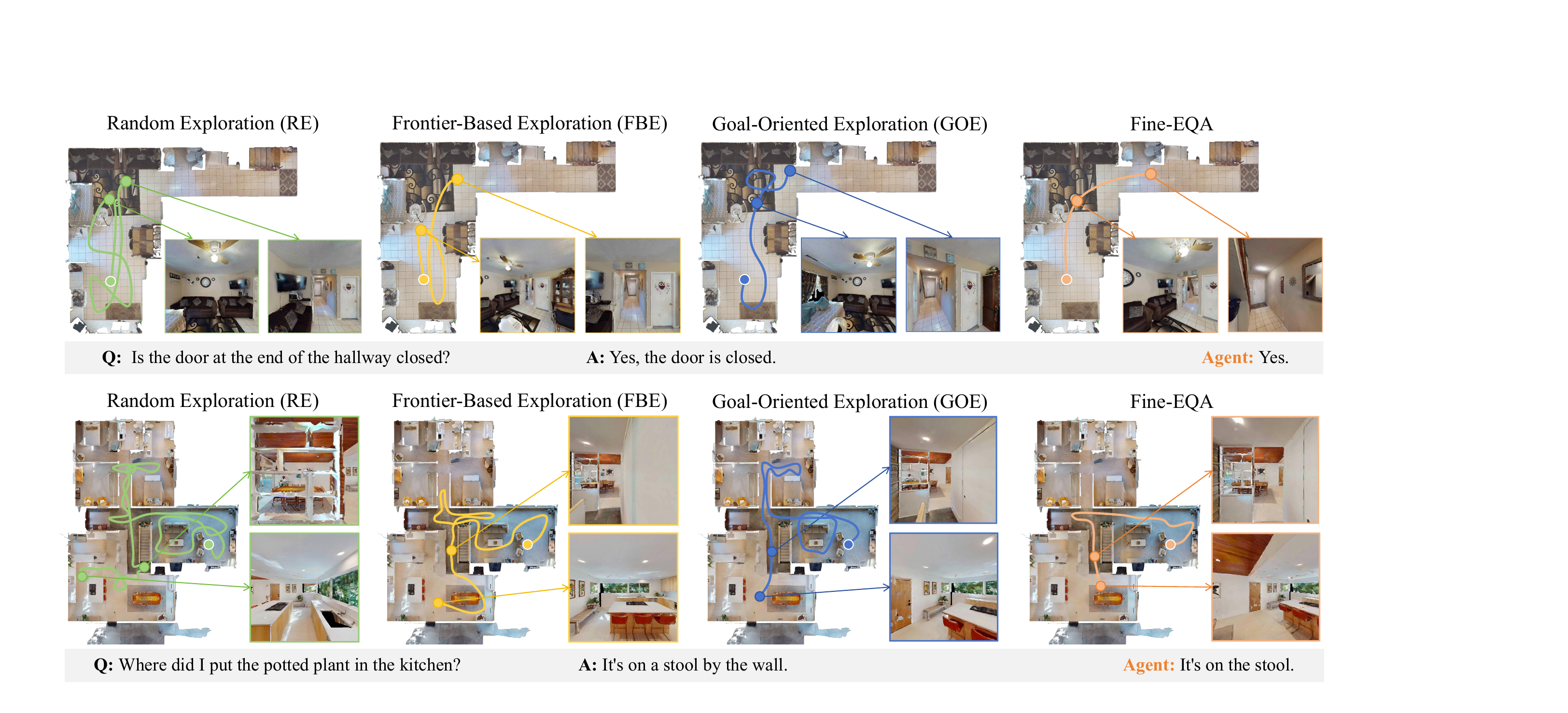}
\vspace{-20pt}
\caption{Visualization of the agent's exploration trajectory under different strategies. While the agent correctly answers the question using all approaches, our Fine-EQA achieves the highest exploration efficiency.}
\vspace{-10pt}
\label{fig:exploration2}
\end{figure*}

\begin{figure*}[t]
\centering
\includegraphics[width=1\textwidth]{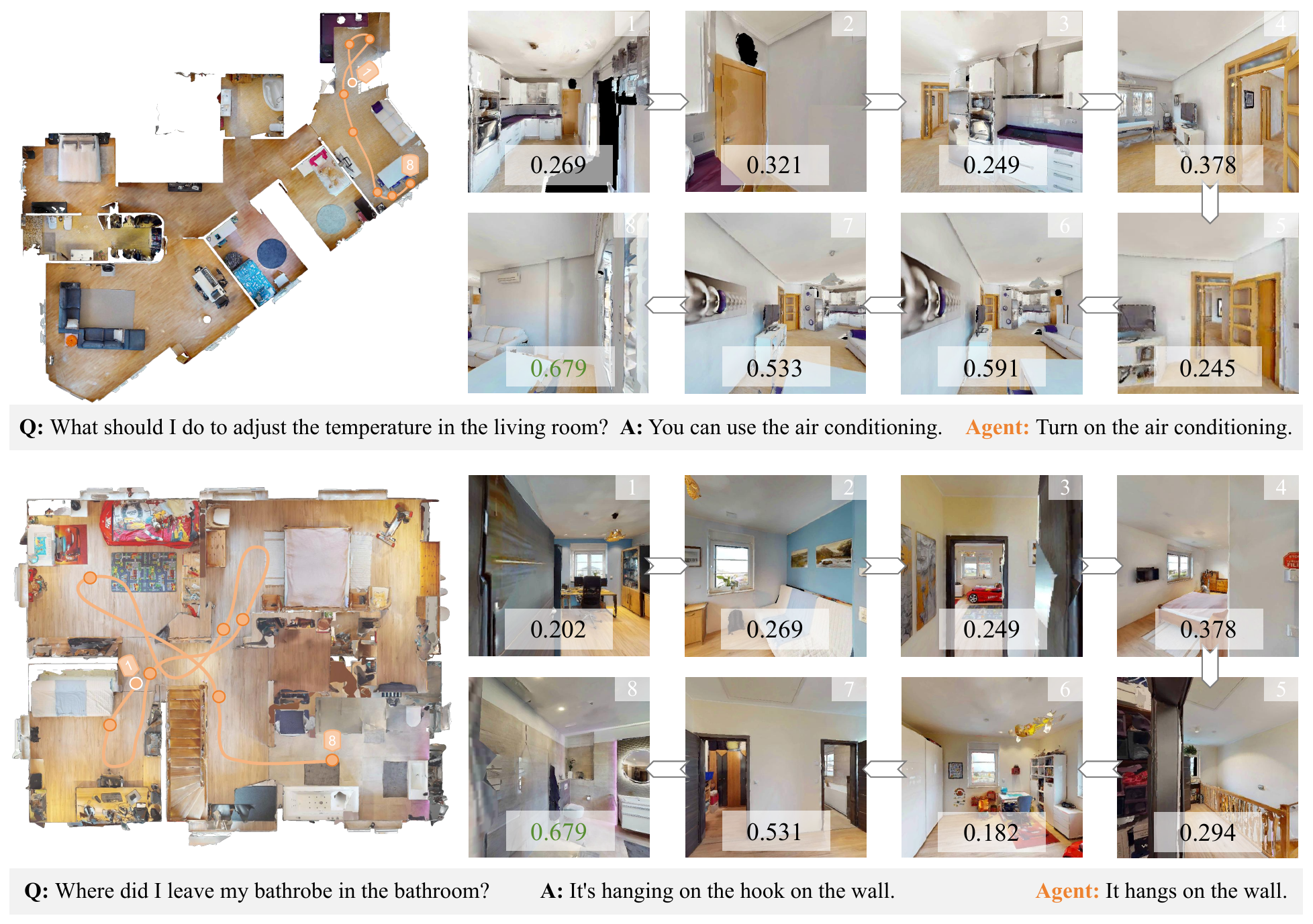}
\caption{Confidence of visual observations at waypoints during the agent’s exploration. The highest confidence is indicated in green in the final frame, demonstrating the reliability of our question reasoning module.}
\vspace{-10pt}
\label{fig:answer2}
\end{figure*}

\begin{figure*}[t]
\centering
\includegraphics[width=1\textwidth]{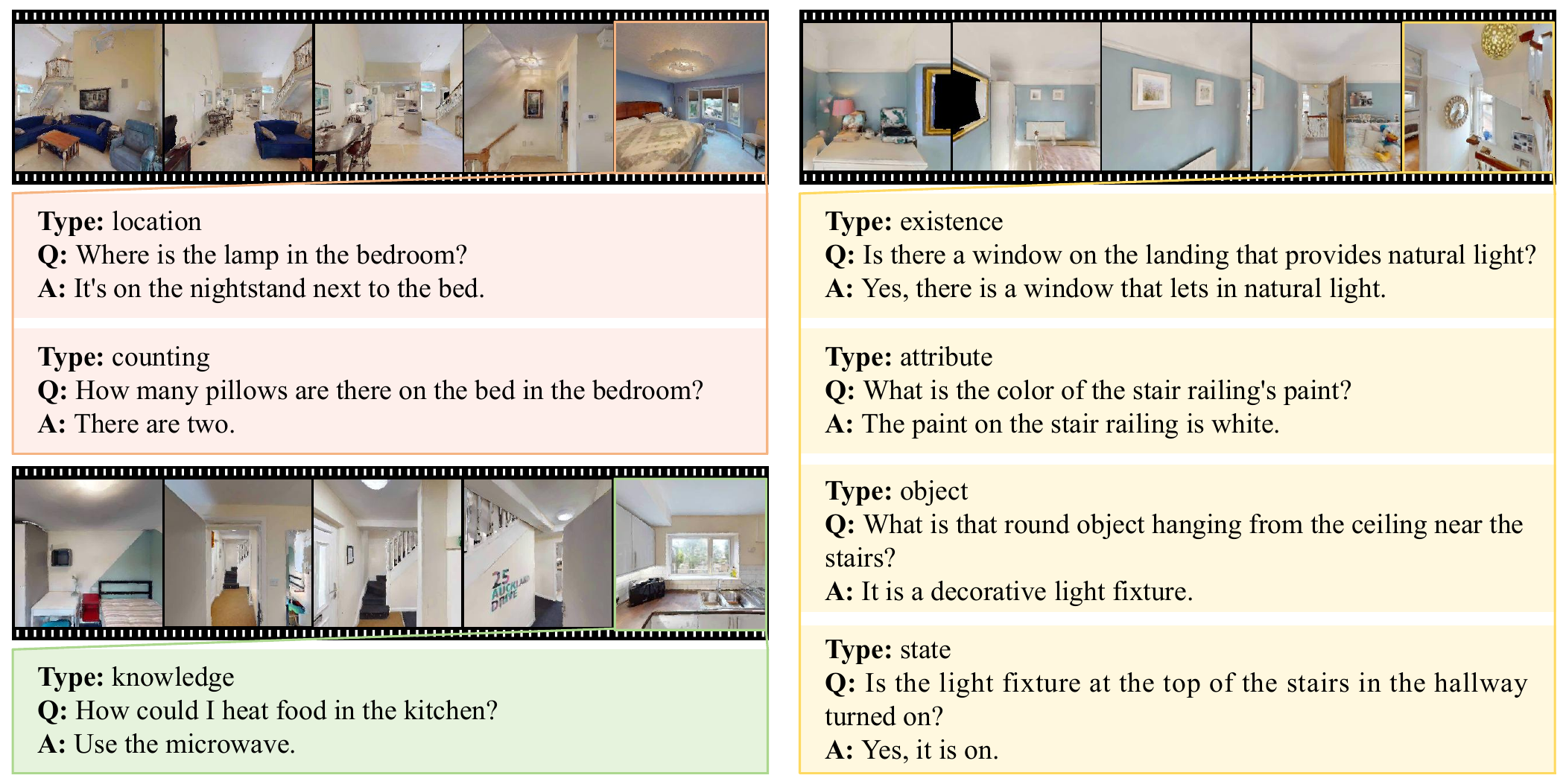}
\vspace{-8pt}
\caption{Examples of different question types from EXPRESS-Bench.}
\label{fig:dataset_examples}
\end{figure*}

\begin{figure}[t]
\centering
\includegraphics[width=0.4\textwidth]{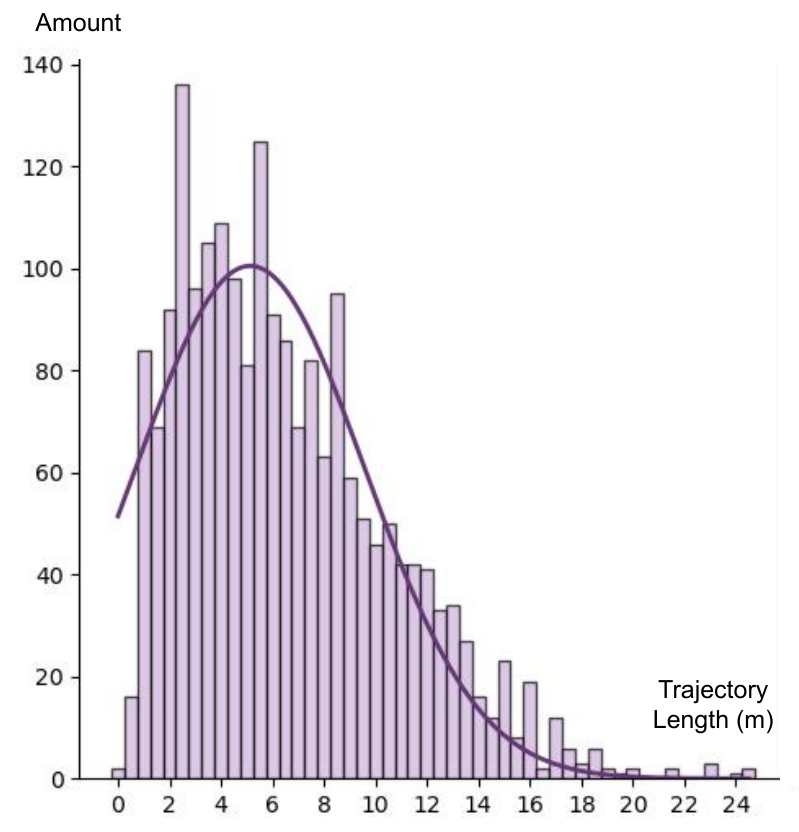}
\vspace{-5pt}
\caption{Distribution of trajectory lengths.}
\label{fig:length}
\end{figure}

\subsection{Performance of Different Problem Types}
\label{subsection:performance}
We categorize the dataset based on question types and evaluate the models' performance across these categories.

Fig.\ref{fig:performance1} presents the $C^*$ scores of all models across different question types. It is evident that human performance significantly surpasses that of all other models.
Overall, the performance gap between models and humans is smallest in the knowledge category, while it is more pronounced in the state, existence, and counting categories. Among the models, Fine-EQA demonstrates strong performance, ranking either the best or second-best in most categories, except for knowledge and existence questions.

Additionally, Fig.\ref{fig:performance2} illustrates the performance of agents with exploration capabilities in terms of the $C$ metric across various question types. After accounting for the grounding of the responses, all agents experience a notable decline in performance. 
While Fine-EQA generally performs well in most categories, its performance on existence-type questions is relatively weaker.

\subsection{Exploration and Answering Effectiveness}
\label{subsection:effectiveness}
We present visualizations of the exploration paths from different agents across additional examples in Fig.\ref{fig:exploration2}. Fine-EQA consistently demonstrated the highest performance.

Fig.\ref{fig:answer2} also presents the confidence scores assigned by VLMs for the agent’s visual observations in two trajectory examples.

\section{More examples of EXPRESS-Bench}
\label{section:dataset_examples}
Fig.\ref{fig:dataset_examples} presents data from seven distinct question types across three tracks. A single trajectory can generate multiple data of different types, all derived from the final frame of the trajectory videos.

We also analyze the distribution of trajectory lengths in the dataset, as shown in Fig.\ref{fig:length}.


\section{Prompt Used}
\label{section:prompt}
We present our data generation prompt ($prompt_1$ \ref{fig:prompt1}), the scoring evaluation prompt ($prompt_2$ \ref{fig:prompt2}), the prompt for determining whether the agent should terminate exploration ($prompt_3$ \ref{fig:prompt4}), and the prompt for answering questions ($prompt_4$ \ref{fig:prompt3}). The design of $prompt_4$ is inspired by ~\cite{majumdar2024openeqa}.

\begin{figure*}[!t] 
\centering
\begin{tcolorbox}[
    colback=gray!10,      
    colframe=black,       
    width=\textwidth,     
    boxrule=0.5pt,        
    arc=4mm,              
    left=2mm, right=2mm,  
    top=2mm, bottom=2mm   
]
\input{prompt1}
\end{tcolorbox}
\vspace{-8pt}
\caption{$prompt_1$ for data generation. }
\label{fig:prompt1}
\end{figure*}

\begin{figure*}[!t] 
\centering
\resizebox{\textwidth}{!}{
\begin{tcolorbox}[
    colback=gray!10,      
    colframe=black,       
    width=\textwidth,     
    boxrule=0.5pt,        
    arc=4mm,              
    left=2mm, right=2mm,  
    top=2mm, bottom=2mm   
]
\input{prompt2}
\end{tcolorbox}}
\vspace{-8pt}
\caption{$prompt_2$ for scoring evaluation. }
\label{fig:prompt2}
\end{figure*}

\begin{figure*}[t] 
\centering
\begin{tcolorbox}[
    colback=gray!10,      
    colframe=black,       
    width=\textwidth,     
    boxrule=0.5pt,        
    arc=4mm,              
    left=2mm, right=2mm,  
    top=2mm, bottom=2mm   
]
\input{prompt4}
\end{tcolorbox}
\vspace{-8pt}
\caption{$prompt_3$ for determining whether the agent should terminate exploration.}
\label{fig:prompt4}
\end{figure*}

\begin{figure*}[!t] 
\centering
\begin{tcolorbox}[
    colback=gray!10,      
    colframe=black,       
    width=\textwidth,     
    boxrule=0.5pt,        
    arc=4mm,              
    left=2mm, right=2mm,  
    top=2mm, bottom=2mm   
]
\input{prompt3}
\end{tcolorbox}
\vspace{-8pt}
\caption{$prompt_4$ for question answering.}
\label{fig:prompt3}
\end{figure*}
\end{document}

%% file: prompt1.tex
You are an expert at generating embodied question answering datasets.

The input is an image. You need to generate questions and corresponding answers based on the image to expand the embodied question answering task dataset. The questions are asked from the perspective of the owner of the house. The robot tasked with these questions needs to navigate around the house, exploring the environment until it captures an observation similar to the input image to gather the information required to answer the question.

Below are examples of different types of questions. Refer to these examples to generate appropriate questions and answers. Your output does not need to cover every type of question; you can generate multiple questions of the same type. The structure of the questions can differ from the examples, but they should fit the respective types.
Make sure the output is formatted the same as the examples. No additional explanatory text is needed. The input image is assumed to be your first-person observation, so your output should avoid phrases like ``in the image".

\vspace{5pt}
type: state\par
question: Did I turn off the light in the living room before I went out?\par
answer: No, the living room light is still on.

\vspace{5pt}
type: knowledge \par
question: What could I do to cool down in the living room?\par
answer: Turn on the air conditioning.

\vspace{5pt}
type: location\par
question: I forgot where I leave my mug in the dining room. Do you see it?\par
answer: It's on the corner of the dining table.

\vspace{5pt}
type: attribute\par
question: Is the floor in the bedroom wooden?\par
answer: Yes, it is made of wood.

\vspace{5pt}
type: counting\par
question: There will be five guests visiting in the afternoon. Are the back cushions on the sofa in the living room enough for them?\par
answer: No. There are only three back cushions on the living room sofa.

\vspace{5pt}
type: existence\par
question: Is there a dressing mirror in the bedroom for me to see how my clothes look?\par
answer: No.

\vspace{5pt}
type: object\par
question: What is the gray object I put on the bed in my bedroom?\par
answer: It's a hat.

%% file: prompt2.tex
You are an AI assistant who will help me to evaluate the response given the question, the correct answer and the scene observed by the robot.

The input includes the Question, the Answer, the Response given by the model and the Image of the environmen. You need to evaluate the alignment between the Response and the Image, as well as between the Response and the Answer, and assign a score for each.

First, assess whether the Response depends on the observed environment Image and assign one of three possible scores [0, 0.5, 1]. If the target object referenced in the Question or the Answer is present in the Image and is described accurately, assign a score of 1. If the object is present but inaccurately described, assign a score of 0.5. If the object does not exist in the Image, meaning the answer is entirely unrelated to the Image and fabricated, assign a score of 0.
Additionally, compare the model's Response with the Answer and Image, assigning a score scale from 1 to 5 based on its accuracy.

Here are some examples illustrating the degree to which response align with the correct answer, accompanied by an explanation of the score provided in parentheses.

\vspace{5pt}
Question: There will be 4 guests. Are there enough chairs around the dining table?\par
Answer: Yes, there are 6 tables.\par
Response: Yes.\par
Your mark: 5 (Correct answer. Giving a specific number is not necessary for this question.)

\vspace{5pt}
Question: What color is the sofa in the living room?\par
Answer: It is light beige.\par
Response: White.\par
Your mark: 4(The output is close to the answer but deviates.)

\vspace{5pt}
Question: Are the curtains in the living room closed?\par
Answer: No, the curtains are partially open.\par
Response: Yes, the curtains are closed.\par
Your mark: 3(The output is close to the answer but deviates because the curtain is not completely closed.)

\vspace{5pt}
Question: Can you tell me where the light switch for the basement is?\par
Answer: It is on the wall near the entrance door.\par
Response: The light switch on the wall near the door.\par
Your mark: 5(The output is completely correct.)

\vspace{5pt}
Question: What could I do if I get cold in the living room?\par
Answer: You can use the blanket on the couch next to the window.\par
Response: You can turn on the fireplace.\par
Your mark: 5(The response is inconsistent with the answer but consistent with common sense, and a fireplace can be observed in the image.)

\vspace{5pt}
Question: Are there any plants in the living room?\par
Answer: Yes, there is a plant near the sofa.\par
Response: No.\par
Your mark: 1(The output is the opposite of the answer.)

\vspace{5pt}
Question: What is the blue item on the bed in the nursery?\par
Answer: It's a baby blanket.\par
Response: It's a coat.\par
Your mark: 2(Object identification error.)

\vspace{5pt}
Your output should consist of exactly two fractions, separated by a comma. No further elaboration is necessary. Please provide the output that fulfills these criteria given the input.

%% file: prompt4.tex
You are an intelligent assistant tasked with determining whether the given image contains sufficient information to answer the provided question.  

The input consists of QUESTION and IMAGE. The QUESTION is what you need to evaluate, while the IMAGE represents the currently observed environment. 

Respond only with ``yes" or ``no" without attempting to answer the question itself.

%% file: prompt3.tex
You are an intelligent question answering agent. I will ask you questions about an indoor space and you must provide an answer.

You will be shown a image that have been collected. Given a user query, you must output `text' to answer to the question asked by the user. No explanatory text is required.

If the query and the image do not provide enough information to properly answer, provide an appropriate guess. Avoid stating uncertainty about answering a question. 
Below are several examples.

\vspace{5pt}
Q: What machine is on top of the stove?\par
A: The microwave.\par
Explanation: Stoves are typically found in kitchens and near microwaves.

\vspace{5pt}
Q: What piece of furniture is in the middle of the bedroom?\par
A: It is a bed.\par
Explanation: Bedrooms almost always contain a bed.

\vspace{5pt}
Q: Is the door open or closed?\par
A: The door is open.\par
Explanation: The door can be in either state, so we just randomly pick one.
